\documentclass{article}

\usepackage{arxiv}

\usepackage[utf8]{inputenc} % allow utf-8 input
\usepackage[T1]{fontenc}    % use 8-bit T1 fonts
\usepackage{hyperref}       % hyperlinks
\usepackage{url}            % simple URL typesetting
\usepackage{booktabs}       % professional-quality tables
\usepackage{amsfonts}       % blackboard math symbols
\usepackage{nicefrac}       % compact symbols for 1/2, etc.
\usepackage{microtype}      % microtypography
\usepackage{lipsum}         % Can be removed after putting your text content
\usepackage{graphicx}
\usepackage{natbib}
\usepackage{doi}
\usepackage{multirow}
\usepackage{booktabs}

\usepackage[flushleft]{threeparttable}
\usepackage{ragged2e}
\usepackage{amsmath, amssymb}
\usepackage{cleveref}       % smart cross-referencing
\usepackage{algorithm}
\usepackage{algpseudocode}
\usepackage{tabularx}
\usepackage{threeparttable}
\usepackage{caption}
\usepackage{epstopdf}
\title{Web3 Meets AI Marketplace: Exploring Opportunities, Analyzing Challenges, and Suggesting Solutions}

\newif\ifuniqueAffiliation
\uniqueAffiliationtrue

\ifuniqueAffiliation % Standard variant of author block
\author{ \href{https://orcid.org/0000-0001-5722-0977}{\includegraphics[scale=0.06]{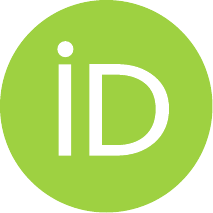}\hspace{1mm}Peihao Li}\thanks{lipeihao.com} \\
	Intellichain Solutions KFT\\
	Honvéd utca 8. 1. em. 2. ajtó\\
	1054 Budapest, Hungary \\
	\texttt{peihao.li@kaust.edu.sa} \\
}
\else
\usepackage{authblk}

\newbox{\orcid}\sbox{\orcid}{\includegraphics[scale=0.06]{orcid.pdf}} 
\author[1]{%
	\href{https://orcid.org/0000-0000-0000-0000}{\usebox{\orcid}\hspace{1mm}David S.~Hippocampus\thanks{\texttt{hippo@cs.cranberry-lemon.edu}}}%
}
\author[1,2]{%
	\href{https://orcid.org/0000-0000-0000-0000}{\usebox{\orcid}\hspace{1mm}Elias D.~Striatum\thanks{\texttt{stariate@ee.mount-sheikh.edu}}}%
}
\affil[1]{Department of Computer Science, Cranberry-Lemon University, Pittsburgh, PA 15213}
\affil[2]{Department of Electrical Engineering, Mount-Sheikh University, Santa Narimana, Levand}
\fi

\renewcommand{\shorttitle}{\textit{arXiv} Template}

\hypersetup{
pdftitle={A template for the arxiv style},
pdfsubject={q-bio.NC, q-bio.QM},
pdfauthor={David S.~Hippocampus, Elias D.~Striatum},
pdfkeywords={First keyword, Second keyword, More},
}

\begin{document}
\maketitle

\begin{abstract}
	Web3 and AI have been among the most discussed fields over the recent years, with substantial hype surrounding each field's potential to transform the world as we know it. However, as the hype settles, it's evident that neither AI nor Web3 can address all challenges independently. Consequently, the intersection of AI and Web3 is gaining increased attention, emerging as a new field with the potential to address the limitations of each. In this article, we will focus on the integration of web3 and the AI marketplace, where AI services and products can be provided in a decentralized manner (DeAI). A comprehensive review is provided by summarizing the opportunities and challenges on this topic. Additionally, we offer analyses and solutions to address these challenges. We've developed a framework that lets users pay with any kind of cryptocurrency to get AI services. Additionally, they can also enjoy AI services for free on our platform by simply locking up their assets temporarily in the protocol. This unique approach is a first in the industry. Before this, offering free AI services in the web3 community wasn't possible. Our solution opens up exciting opportunities for the AI marketplace in the web3 space to grow and be widely adopted.
\end{abstract}

% keywords can be removed

\section{Introduction}
Artificial Intelligence (AI) and Web3, as two big players in the world of tech, have caught everyone's attention in recent years. Think of AI as the capability of computers to mimic human intelligence, helping us do things like assisting programming, driving cars, or even diagnosing illnesses. On the other hand, Web3 is like a new version of the internet, where instead of a few big central entities holding all the power and data, it's distributed, making everything more transparent and fair.

\subsection{Web3 at a glance}
Web3, often termed as the "new internet", is the next phase in the progression of the World Wide Web (WWW). If web1.0 was about static web pages and read-only content, and web2.0 brought interactivity, social media, and user-generated content, then web3 is about decentralized and trustless protocols and technologies. It moves away from centralized control and ownership, as seen with big tech companies today.

At its core, web3 is primarily built on decentralized blockchain technology. It emphasizes user control over personal data, trustless interactions (meaning you don't need third-parties to trust each other), and direct peer-to-peer exchanges of value. In some situations, web3 can perform tasks more effectively than traditional web applications, and in certain cases, it can achieve what traditional platforms cannot. Below, we summarize some of the hottest sectors in web3:
\begin{enumerate}
    \item \textbf{Decentralized Marketplaces:} Peer-to-peer marketplaces where users can transact directly without middlemen. This applies to both goods and services. Decentralized finance (DeFi) stands out as one of the most significant sectors in decentralized marketplaces which aims to recreate traditional financial systems, such as loans, savings, insurance, and more, in a decentralized manner using smart contracts on blockchains. Unlike traditional finance, DeFi operates in a fixed and transparent manner, and there is no room for hidden activities behind the scenes in such financial products. 

    \item \textbf{Borderless Transactions:} Traditional financial systems often impose high fees and delays on international transfers. Cryptocurrencies, such as Ripple \cite{schwartz2014ripple}, allow for almost instantaneous global transactions with minimal fees amounting to mere cents.
    
    \item \textbf{Digital Authenticity:}  Traditional digital files can be copied endlessly, making it hard to identify the "original." Non-Fungible Tokens (NFTs), on the other hand, provide a unique stamp of authenticity that can't be duplicated. Every sale or transfer is transparently recorded, ensuring true ownership and history. This means artists and creators can sell their work digitally, knowing there's a verifiable "original" out there. NFTs have gained massive traction in art, collectibles, and even real estate in the virtual space.
    
    \item \textbf{Decentralized Decision Makings:} Traditional organizations have a hierarchical structure where decisions often come from the top. DAOs (Decentralized Autonomous Organizations) operate on consensus mechanisms, allowing all members to have a say. Without a central authority, decisions can be made transparently and collectively, ensuring every stakeholder's voice is heard and reducing the risks of centralized corruption or biases.
    
    \item \textbf{Decentralized Web Infrastructure:} This includes decentralized file storage solutions with platforms like Filecoin \cite{filecoin2017} and IPFS (InterPlanetary File System) \cite{DBLP:journals/corr/Benet14}, GPU service systems \cite{rendernetwork2023}, blockchain oracles \cite{oracle} and more, which ensure that the foundational aspects of the internet are distributed and not controlled by any single entity.
    
\end{enumerate}

\subsection{AI's explosive growth}
Previously, notable AI milestones, such as AlphaGo \cite{silver2016go}, which defeated a world champion Go player in 2016, were celebrated but quickly faded from the public’s consciousness. Later on, the global financial market witnessed the explosive growth of generative AI (gen AI) tools in 2022, capable of producing various types of content, including text, imagery, audio, and synthetic data. Among these tools, OpenAI's GPT implementation stands out as the AI-powered chatbot that took the world by storm. Driven by incredible market reactions, it is estimated that ChatGPT reached 100 million monthly active users in January 2023 \cite{popularchatgpt}. In comparison, TikTok took nine months to achieve 100 million users, while Instagram took 2.5 years. Experiments show that ChatGPT significantly increased productivity, with the average time taken reduced by 40\% and output quality improving by 18\% \cite{2023productivity}.

Although generative AI caught the public's attention in 2023, AI is generally considered to include other main sectors such as supervised machine learning, unsupervised machine learning, and reinforcement learning. The marketplace of AI as a whole is expanding rapidly. The global AI solution market is projected to attain 301.2 billion USD by 2028, with a compound annual growth rate (CAGR) of 29.4\%. Specifically, the unsupervised machine learning sector is forecasted to reach 15.6 billion USD by 2028, expanding at a CAGR of 25.1\%. Furthermore, AI solutions deployed in public cloud environments are expected to triple the figures of private cloud implementations during the same period \cite{mindcommerce2023ai}. Regarding regions, North America generated more than 36.84\% of the market share in 2022. The Asia Pacific market is expected to expand at the highest CAGR of 20.3\% from 2023 to 2032 \cite{aimarket2023}.

While AI and Web3 have their distinct advantages both in utilities and marketplace, a growing area of interest lies in their combined potential. The idea is simple: what if we could merge the decentralized approach of Web3 with the capabilities of AI? This combination could lead to more powerful AI systems that are also more accessible to everyone. Consider the possibilities, such as using Web3's decentralized structure to train AI models, or making AI tools available to a wider audience through Web3 platforms.

In this article, our focus is on understanding how AI and Web3 can be merged, examining the potential opportunities and addressing the challenges. We will dive into topics currently of great interest in this field. Our aim is to provide a clear and informed perspective on the research intersections between these two transformative technologies.

\section{Opportunities and Challenges}
In this paper, we are particularly interested in the integration of the web3 infrastructure with the AI marketplace, an area where web3 enhances AI's product performance in the marketplace. While there are various ways that AI and web3 can complement each other, such as using AI to produce NFTs or employing AI as an autopilot for code writing, our focus is on the opportunities in the current commercial domain. However, we approach this with a broader vision, digging into the potential of both AI and web3. In following subsections, we will explore the opportunities and challenges of such an integration.

\subsection{Why is web3 more accessible?}

According to the statistics of the world bank in 2021 \cite{worldbank2021}, credit card ownership seemed to align with a nation's development. Canada led with 82.7\%, followed closely by developed countries like Israel, Iceland, and Japan. The USA stood at 66.7\%. Many European nations reported over 50\% ownership. In contrast, many African and South Asian countries, such as Nigeria and Pakistan, recorded less than 2\%. The trend suggests that developed countries have a higher percentage of credit card holders compared to less economically advanced nations. As a result, people in generally less-developed countries have limited access to paid AI services due to a lack of payment methods. Furthermore, teenagers under the age of 15 typically have limited access to credit cards. Consequently, they have restricted access to AI services and support. If they wish to use these services, they often have to rely on their parents' cards, creating transactional friction and making AI less accessible to the general public.

Web3 and crypto, however, offer a much simpler payment process. Taking Ethereum as an example, since anyone can create an Ethereum wallet easily by setting up everything on their mobile with a few clicks, it's more user-friendly than traditional banking and centralized payment methods. And having an Ethereum wallet and ether means you have access to the web3 world, so almost everyone can access web3 if they buy any type of crypto. Moreover, people are starting to accept crypto as salaries since they don't have to use an international bank to receive it \cite{cryptosalary}. Famous comments, such as one from Vitalik in 2020, mention workers in Africa accepting ETH as payment. A broader and simpler access through web3 could make AI services more available and exciting as an industry.

\subsection{Web3 infrastructure as an advantage}
Any public chain requires a consensus mechanism to update the global states in a distributed network system, with Proof of Work (PoW) being the most commonly used \cite{Consensus1, Consensus2}. The consensus mechanism is often referred to as crypto mining, which involves creating and adding new blocks to a blockchain network using various consensus methods based on different resources (like mining rigs, staked tokens, etc.). In the PoW consensus mechanism, miners compete to produce the next valid block by being the first to solve a cryptographic puzzle, thereby earning a reward for their efforts. In the marketplace, as a public chain gains popularity, its crypto miners receive increased rewards. This attracts more miners, or in other words, more computing power, to join the chain. Consequently, the total hash power of the chain continues to rise over time. Notably, prominent blockchain projects in the crypto industry, such as Bitcoin (BTC) and Ethereum (ETH), have used the PoW consensus mechanism for years. According to Bitcoin energy consumption analysis \cite{BTCConsumption, BTCConsumption2}, the annual electricity consumption of Bitcoin mining surpassed that of the United Arab Emirates (119.45 TWh) in 2021 and Sweden (131.79 TWh) in 2022. Most of this energy is dedicated to solving cryptographic puzzles.

While this process achieves trustless consensus, it doesn't offer practical benefits beyond producing block hashes that, in Bitcoin's case, have a certain number of zeros at the beginning \cite{bitcoinarticle}. Consequently, the absence of a theoretical limit on energy consumption for the PoW mechanism has raised global concerns. This led to the exploration of alternative consensus mechanisms, like Proof of Stake (PoS), and changes in institutional policies. For example, in 2021, Tesla announced it would no longer accept BTC due to climate concerns \cite{swissgovernment2021}. In 2022, Ethereum transitioned from the energy-intensive Proof of Work (PoW) mechanism to the more efficient Proof of Stake (PoS) in response to environmental concerns. This shift resulted in a significant reduction in energy demand, with decreases ranging from 99.84\% to 99.9996\% \cite{POWReduce}. This reduction in Ethereum's energy consumption is comparable to the electricity needs of countries like Ireland or even Austria, marking a notable step towards environmental sustainability. However, this change also left a large amount of unused hashrate, equivalent to 1,126,674 GH/s \cite{etherscan2023hashrate}, without a specific use. The advancement of computing resources in crypto mining isn't the only type of resource in web3. Linear or exponential growth in the infrastructure supporting specific consensus algorithms has been widespread in most mainstream web3 utility projects. For instance, in cloud storage, the capacity of decentralized storage has surged from just a few million TB to nearly a hundred EiB \footnote{1 EiB $=$ 1,152,921,504.6068 GB}, as per reports and statistics \cite{filfox2023, winwin2022}. Furthermore, the cost of decentralized storage is on average \$0.19 per month, which is much cheaper than centralized solutions such as Dropbox.

Meanwhile, as artificial intelligence (AI) becomes integrated into various sectors of the economy, there's a rapidly growing demand for computational resources to power this machine intelligence. Training a model like ChatGPT can cost over \$5 million, and the initial operation of the ChatGPT demo ran OpenAI an approximate \$100,000 daily, even before its current usage surged \cite{2023cryptominersAI}. Midjourney, a service that provides high-quality images, operates with more than 9,000 GPU cards, contributing to its operational costs. Given the vast number of neural parameters and extensive GPU hours involved, the high computational demands of model optimization pose significant challenges for academic researchers and small-scale enterprises. This limits the broader adoption and use of artificial intelligence technologies.

It is, therefore, unsurprising that an increasing number of crypto miners are exploring ways to use their existing computational infrastructures to advance AI. They are redirecting computational resources, which were previously focused on mining, toward machine learning and other high-performance computing (HPC) applications, such as the Internet of Things (IoT) and data services \cite{fetchai, oceanprotocol}. Another example is provided by Hive Blockchain, which is shifting its long-term HPC strategy from Ethereum mining to applications like artificial intelligence, rendering, and video transcoding, contributing to their total annual revenue generation of approximately \$102 million. Miners can also opt to employ these resources for processes on decentralized blockchain networks.

\subsection{Challenges of Merging AI and Web3 Infrastructure}

While there are significant opportunities in both marketplaces, we have identified major challenges that prevent the development of standout applications. We will analyze these difficulties from both market and technical perspectives to better inform potential solutions for integrating web3 with the AI marketplace.
\begin{enumerate}
    \item \textbf{Blockchain resources are inherently costly:} When it comes to the blockchain consensus infrastructure, resources are, by design, typically expensive. The FLP impossibility theorem states that in an asynchronous distributed system, where at least one process can fail, it's impossible to design a consensus algorithm that simultaneously guarantees both safety and liveness \cite{FLPimpossible}. This is a primary reason most blockchain systems adopt synchronous or partially synchronous\footnote{There's an assumption that a time upper bound exists for message delivery and block production; however, this bound may be unknown or subject to change during system upgrades.} consensus mechanisms such as bitcoin or ethereum. However, such systems often have substantial storage and bandwidth costs, especially since they store $n$ replicas of the global states. It's therefore essential for the protocol to maintain only the necessary states, minimizing storage requirements. Given the rapid development of the AI marketplace, embedding the entire system into a layer-1 (L1) blockchain solution\footnote{A Layer-1 (L1) blockchain represents the fundamental tier of a blockchain network, comprising the base protocol that oversees the consensus mechanism, transaction processing, and data storage. This layer delivers the core functionality of the blockchain system and supplies the infrastructure for crafting additional layers or applications atop it.} might not be the most efficient strategy \cite{PoLe, coinAI}. Such systems generally uphold a consistent block production rate\footnote{The block production rate in a blockchain denotes the rate at which new blocks are generated and appended to the blockchain. For instance, the TRON (TRX) network boasts a rapid block production rate, with a fresh block produced every 3 seconds.}, thereby ensuring a consistent transaction throughput capacity. However, in the case of decentralized AI marketplace, the workload can vary dynamically based on market supply and demand. There might be instances where the system witnesses inactivity due to an absence of incoming training tasks, resulting in most nodes becoming stale without a continuous reward stream. In this setup, the primary goal is to orchestrate AI market activities in the network, with transaction validation serving as a secondary role. A well-constructed framework should:
    \begin{itemize}
        \item address these aspects by dynamically adjusting system workload based on the influx of jobs and tasks
        \item enable seamless system upgrades over time
        \item ensure the ease of use and security for users’ assets
    \end{itemize}
    Unfortunately, such a system is currently lacking in the industry.
    \item \textbf{Payment frictions in AI service subscriptions:} Even with the assumption that cryptocurrencies offer easier access and operation, the business revenue model for various AI services remains a challenge. While customers are willing to pay for specific tasks, they resist being charged repeatedly when switching between services—a common occurrence in traditional AI businesses. For instance, if you purchase a ChatGPT premium for access to GPT-4 and additional features, you'd still have to pay separately for a Midjourney premium should you need its services. Consequently, customers wanting to use a broad array of AI services could face hundreds of dollars in monthly subscription fees. Even within the same company or network, users don't want to be charged each time they order tasks, as seen with the GPU tasks pricing model in the render network \cite{rendernetwork2023}. Exploring how web3 solutions can enhance the user experience regarding subscription practices is of significant interest.
    \item \textbf{Integrating multiple parties:} In the traditional AI business model, there is a direct value exchange between two main parties: the customers and the service providers. Similarly, in most blockchain models, there are only two primary participants: the crypto users who send the transactions and the crypto miners who validate those transactions. As evident, both traditional AI products and web3 communities involve only two major parties. While web3 infrastructure has the potential to broaden the accessibility of AI and offer better market rates, its integration introduces additional participants into the network, thereby increasing complexity. In general, there are at least three parties involved: the customer, the miner providing computing power and storage, and the product designers who contribute the foundational building blocks for various AI services. Developing a system framework and reward models that benefit all three major parties poses significant challenges.
    \item \textbf{Securities:} Web3 emphasizes decentralization. However, distributed systems are inherently unstable and insecure. In designing the system/framework we describe, we must account for a significant number of nodes being faulty or malicious up to a certain percentage. All blockchain systems employ mechanisms to prevent attackers from initiating various types of attacks. Put simply, attacking the system should be more costly financially for the attacker than the total potential reward they might gain from the attack. Consequently, different blockchain systems implement their own consensus mechanisms to prevent attackers from forging and tampering with data and states \cite{Consensus1, Consensus2,consensuscomparison}, with Proof of Work (PoW) being the most widely adopted. 
    
    The consensus mechanism for integrating web3 and AI requires a novel design, as proving service provision can be quite tricky. This mechanism must account for various participant roles. Primarily, it needs to ensure that service providers are executing their tasks both honestly and diligently. If service providers conduct denial of service or provide low quality service to too many customers, the system should either forfeit some of their rewards or, at a minimum, impact their reputation. This will alert future customers to be wary of these specific providers. Moreover, if a reward forfeiture or reputation system is part of the consensus mechanism, there must also be a safeguard against customers providing unjust or malicious reviews. Without a robust protocol, genuine service providers could become targets of sybil attacks. Lastly, nodes responsible for maintaining global state records must be given sufficient cryptoeconomic incentives to act both honestly and diligently, given their crucial role in ensuring system security. 
    
    To the best of our knowledge, such protocols addressing all the challenges mentioned above are currently lacking in the research field, and we don't see many implementations in the industry field, other than fetch.AI and singularityNET \cite{fetchai,backgroundPaper1} which partially addressed the challenges. While the potential market size and areas of opportunity  can be tremendous, we believe that the following challenges, as summarized from previous discussions, must be addressed to succeed in the large-scale commercialization of the AI marketplace integrated with web3 infrastructure.
    \begin{itemize}
        \item \textbf{Protocol capacity and scalability:} The system/platform should be capable of coordinating clients, miners, and AI product development, and it should empower self-governance to initiate, process, and finalize services. The volume of transactions—including client orders, reward claims, and network management—will largely be determined by the customer base and the size of the AI marketplace within the network. The computational power needed to maintain the system's global states should be possible to analyze theoretically. Additionally, the protocol should consider certain commercial factors, integrating "free" features that are specific to the blockchain industry, such as the inflation model, to enhance its appeal to potential customers.
        \item \textbf{Protocol securities:} Given the nature of the AI marketplace, it's impractical for a central ledger to check on every transaction, such as AI services, to ensure they're executed honestly—both in theory and practice. AI services require substantial computation, and given that many incorporate randomness and don't lead to a single definitive outcome, it's theoretically challenging for other nodes to determine if a single node is functioning accurately. Thus, a protocol safeguarded by cryptoeconomics—where attacking the system costs more than complying with it—is preferable. To launch an attack, one would typically need more tokens than the counterparties, which can often be financially impossible. In other words, without significant potential rewards, there's little motivation to compromise the protocol. Systems must be intricately constructed to prevent the potential rewards from being so attractive that the system's design itself becomes a target for malicious activities. When attackers recognize that their attacks will be easily corrected by the system, they have tiny incentive to proceed.
    \end{itemize}
\end{enumerate}

\section{Analyzing Solutions}
In this section, we will propose general guidelines and possible solutions by analyzing the pros and cons in different architecture design and implementations. Our primary focus is on two different aspects: the technical aspects and the economic aspects. In the technical aspects, we will focus primarily on system security and efficiency, and in the economic aspect, we will focus on customer experience and diversity in service subscriptions.
\subsection{L1, L2 or L1-L2 architecture?}
Blockchains based solely on L1 have their own mechanisms of producing blocks, while blockchains comprising both L1 and L2 \footnote{A Layer 2 (L2) in blockchain refers to a secondary protocol or framework built on top of an existing blockchain, primarily aiming to enhance the network's scalability, efficiency, and transaction throughput. Layer 2 solutions leverage the security and decentralization of the underlying blockchain (Layer 1), while offloading a portion of the computational workload to a separate network or system. This enables faster and cheaper transactions, as well as more complex operations, without burdening the base layer. Examples of Layer 2 solutions include state channels, sidechains, and rollups.} place most of the utility/core infrastructures on L1 for efficiency and move most token logistics and value storage to the L2 layer. This setup often relies on many other well-known blockchain ecosystems to serve a wider range of customers and investors. L2 solutions typically embed their program logic and database into smart contracts within mainstream ecosystems. These projects integrate their logic entirely into smart contracts, complemented by a frontend framework connected to the backend contracts.

\begin{table*}[hbt!]
\begin{threeparttable}
\caption{Performance comparison of L1, L2 and L1-L2 architecture}
\label{tab:2}
\setlength\tabcolsep{0pt} % make LaTeX figure out intercolumn spacing

\begin{tabular*}{\textwidth}{@{\extracolsep{\fill}} ll ccccc}
\toprule
     Architecture & Mainnet & 
     \multicolumn{5}{c}{Performance} \\ 
\cmidrule{3-7}
     & & fees on-chain$^{a}$ & fees off-chain$^{a}$ & supported tokens& stablecoin integration?& upgradability \\
\midrule
     L1& - & $\epsilon^{b}$ & 0 & local &no &difficult \\
      
      \addlinespace
         L2 & Ethereum & 0.0004 units & 0& ETH\&ERC20 &yes&medium\\
 & BSC$^{c}$& 0.000075 units & 0 & BNB\&BEP-20 & yes& medium\\
      & Tron & 0.027 units &0 & Tron\&TRC20 & yes&medium \\
      \addlinespace
L1-L2& Ethereum & 0.0004 units & $\epsilon$ & ETH\&ERC20& yes &easy\\
      & BSC & 0.000075 units& $\epsilon$ & BNB\&BEP-20 & yes& easy\\
      & Tron & 0.027 units & $\epsilon$ &Tron\&TRC20 & yes &easy\\

\bottomrule
\end{tabular*}

\smallskip
\scriptsize
\begin{tablenotes}
\RaggedRight
\item[a] For the public mainnet, data is fetched from the respective blockchain explorer.
\item[b] $\epsilon$ can be either zero or close to zero, depending on the protocol specifications.
\item[c] BSC refers to the Binance Smart Chain.
\end{tablenotes}
\end{threeparttable}
\end{table*}

Table. \ref{tab:2} summarized the performance matrix of different architecture design. L1 ecosystems typically have their own databases and block production mechanisms. All transactions and associated state changes occur on-chain, using their local utility coins/tokens. Transaction fees $\epsilon$ can be set to very small values, as seen in the Tron network \cite{tron2018whitepaper}. However, once initiated, such blockchains can't easily be halted, and upgrades to core functions can be challenging. Such upgrades often necessitate a hard fork by miners or validators, which requires extensive communication between various parties to adopt a new protocol at a predetermined block height \cite{hardfork}. In our effort to integrate web3 infrastructure with the AI marketplace, we need a setup that allows for ongoing system updates and feature additions without disrupting the network's assets or user experience. Building everything on L1 may not be the optimal solution. 

L2 solutions, on the other hand, place all their core logic on a specific public mainnet, eliminating off-chain costs. All activities occur on-chain through contract calls to the mainnet. Assisted by oracles \cite{oracle}, L2 ecosystems span a wide range of areas including DeFi, Gaming, and NFT Marketplaces. Upgrades in L2 are typically handled using the upgradable contract paradigm, where contract updates are achieved by redirecting the proxy contract pointer \cite{salehi2022immutable}. However, given that AI models and product upgrades cannot be fully migrated on-chain, this architecture is not suitable in the given context.

To effectively harness the potential of this decentralized network for web3 and AI marketplace merging, a two-layer L1-L2 architecture is introduced. The on-chain component ($\mathrm{SC}$) records the value flow within the network, while the off-chain component ($\mathrm{exec}$) comprises a set of protocols operating on the distributed network where utilities are executed. By seamlessly integrating the on-chain functionality with the diverse off-chain services provided, the system can achieve the robustness and upgradability that traditional Layer 1 solutions often lack. In the L1-L2 design, protocols and infrastructures mainly operate off-chain within the decentralized network, while token utilities like transfer and withdrawal function on Layer 2 of mainstream blockchains such as BSC or Polygon. This configuration enables the system to regularly update with new features and utilities, all while preserving the network's assets and the user experience. In the AI marketplace, the core module can be designed in L1 to ensure easy upgradability. Participants' databases can be distributed between L1 and L2 by placing their assets in L2 and conducting transactions in L1, thereby increasing efficiency and reducing costs. Protocols such as Chainlink and Proof of Training (POT) \cite{chainlink2017,li2023proof} also adopt the L1-L2 architecture.

\subsection{Economic model analysis: charged vs. uncharged approaches}
Most existing decentralized AI products attempt to collect micropayments for every user request at their own dedicated rates. As a result, these platforms require users to continually purchase cryptocurrency to pay for services and bandwidth. This suggests that the services might not be readily available for the general public to access for free via their browsers. Meanwhile, the quality of services varies widely, ranging from large-scale enterprises offering high-quality services to home computer and GPU providers renting out resources with slow internet connections. Users often remain unaware of the quality of the services they are using, yet they are continuously charged. All of these create transaction frictions and prevent large scale commercialization and adoption of the decentralized AI apps. All of these factors create transaction frictions and prevent large-scale commercialization and adoption of the decentralized AI apps. In the following subsections, we will provide solutions for both charged and uncharged scenarios.
\subsubsection{Uncharged approaches for decentralized AI services}
When trying to design an uncharged platform for decentralized AI services like the "free" YouTube or Gmail in traditional internet, we need to keep in mind that there is no such thing as a free lunch. So, who is actually paying for the AI services and bandwidth provided by the network miners? Existing decentralized solutions all rely on one-time or monthly micropayments, creating transactional friction that discourages adoption. In practice, we typically see strong consumer resistance to micropayments in favor of no fees, flat fees, or one-time payments \cite{friction}. Therefore, to build such approaches, we need to solve the problem of guaranteeing "free" and high-quality services to users while ensuring that network miners are rewarded as they provide an increasing amount of services.

The approach we can take is to draw inspiration from the inflation model of the EOS storage design \cite{eosio2018technical}. In this model, there is a certain percentage of annual inflation on the total coin/token supply of the ecosystem to ensure that miners get paid. Meanwhile the clients will need to lock the platform related coin/token into smart contracts in order to gain allowance of job requests. Service providers\footnote{\textbf{Service providers} and \textbf{miners} can be used interchangeably in this paper's context.
} collectively provide the computational power and AI service capacity to those requests. 
For users to access AI services, they must stake their tokens in the smart contract designated for AI services. Think of this staking process as making a fully refundable security deposit. Users can retrieve their tokens by releasing the service providers from the obligation to provide further AI services to them. This mechanism of staking/locking tokens from the client side will prevent all forms of Sybil attacks, which could flood the system with unlimited requests, halting the system indefinitely. Clients can only secure more service capacity by pledging more tokens to the network compared to other clients.

\paragraph{The system's robustness in general}
There are several major aspects to consider when designing a staking-and-use mechanism like this. First, we must ensure that the miners are motivated to provide honest and high-quality services to the clients. Not only should they possess adequate facilities and resources, such as substantial computational power and network bandwidth, but they should also have associated reputation records. These records would include scores given by clients for their services and the amount of stake they have locked, representing their commitment to the system's overall ecosystem. The system should also keep a record of clients' reviews. If a client continuously gives malicious reviews that deviate significantly from other clients' feedback, the system should implement appropriate penalty mechanisms for such behavior.

\paragraph{The miner's perspectives}
When a client sends a request, the system forwards it to a specific service provider. Assuming the service provider is designed to handle many requests simultaneously, there may be times when the number of incoming requests exceeds its processing capacity. In such instances, requests are queued. The service provider would then prioritize these requests based on the number of tokens each client has staked. We consider that using a weighted round-robin (WRR) scheduling \cite{Weightedrobin} to ensure more predictable and fair access, while still respecting the proportional stake. 

Let's consider that at a specific time $t$, we have a miner $k$ receiving requests from $N_t$ clients, where each client sends a specific number of requests denoted as $(n_1^k, n_2^k, \cdots, n_{N_t}^k)$. Each client has staked tokens in the amounts $(s_1, s_2, \cdots, s_{N_t})$. We can determine the corresponding weight of each client using:
\begin{equation}
    w_r = \lfloor \frac{s_r}{s_{\text{min}}} \rfloor
\end{equation}
where $s_{\text{min}} = \text{min}(s_1, s_2, \cdots, s_{N_t})$
 
Once the weight is determined, miner $k$ will stop accepting further requests by setting the $\texttt{status}$ variable to $\texttt{busy}$. This signals the coordinator nodes to stop forwarding more requests to miner $k$. Let $w_{\text{max}} = \text{max}(w_1, w_2, \ldots, w_{N_t})$. With interleaved WRR, miner $k$ would require $w_{\text{max}}$ rounds to process all of the requests. In each round, one request from each client is processed. Assume there are $N_1$ clients with only one request. Then, in the first round, there are $N_t$ requests to be processed, and in the second round, there are $N_t - N_1$ requests. This procedure continues until all requests have been iteratively processed. Afterward, the $\texttt{status}$ variable of miner $k$ is reset to $\texttt{ready}$. 

Once a job request is processed, the global ledger receives a signed message from the service provider indicating successful completion, and the client obtains the AI service output from the miner. The client can then review the service provided by the miner by sending a signed message to the global ledger that reflects the quality of the service. While proof-of-reputation (POR) is generally used in existing literature as a method to produce blocks and validate transactions \cite{proofofreputation}, we employ the reputation system as a reference for both the global ledger and clients. This may be a desired input for certain utility functions and protocols.
Given ratings Good (G): $1$, Fair (F): $0$, and Bad (B): $-1$, the cumulative reputation for miner (service provider). $R_k(t)$ is computed as:
\begin{equation}
 R_k(t) = 100 \times \frac{1}{1 + e^{-\theta \sum_{i=1}^{N} c_i}}   
\end{equation}
where $c_i$ represents the latest reputation score given by client $i$ and $\theta$ adjusts the sensitivity of the score. This logistic function maps any real number to the range [0, 100], ensuring a bounded and smooth reputation score.

\paragraph{Rewards for miners} As we have discussed the inflation model, we need to dive into how these rewards can be distributed to individual miners. In traditional inflation models, such as the Solana ecosystem \cite{yakovenko2022solana}, rewards are proportionally distributed to validators based on the volume of their staked tokens. In our case, however, the reward system is slightly more complex. Miners are rewarded not only for their computation but also, and more significantly, for the quality of the services they provide. This is determined by their reputation, service provision logs, and corresponding work volume calculated by the global ledger. We suggest that miners be rewarded based on their contribution, denoted as $C_k$ for the miner indexed $k$, during a certain time span between $t_1$ and $t_2$, as calculated by the following formula:
\begin{equation}
    C_k = \sum_{t=t_1}^{t_2} \sum_{j=1}^{M} N_{\text{processed},t,j} \times W_{\text{service},t,j}
\end{equation}
where \( N_{\text{processed},t,j} \) denotes the number of requests processed for the \( j^{\text{th}} \) service at a specific time \( t \). $t_1$ corresponds to the time of the last reward distribution, and $t_2$ represents the time of the next reward distribution. Meanwhile, \( W_{\text{service},t,j} \) signifies the weight associated with the \( j^{\text{th}} \) service's processed requests at that same time. For instance, the output of text-related services generally has a lower weight than that of image-related services due to its computational and memory requirements. Across the entire interval, we're considering \( M \) distinct services in the marketplace, while any new services can be added with weights determined by the community DAO.
\paragraph{The client's perspectives}
The system enhances user-friendliness on the client side by minimizing the necessary work and associated fees. Typically, clients can employ built-in third-party tools or APIs, such as Metamask and TrustWallet, to initiate processes. By staking any cryptocurrency recognized by the L1-L2 network, clients gain access to all AI services available on the platform with just a few clicks. The only cost is the initial staking transaction fee on L2, which can be less than one dollar on public chains like BSC or Polygon. Clients also have the option to rate the services upon receiving content from service providers, submitting their ratings to the coordinator nodes. Furthermore, to prevent DDoS attacks, coordinator nodes can offer the service without any charges but may set a request threshold for each client.

The platform also makes it possible to stake any crypto assets by leveraging the L1-L2 infrastructure. This means that as long as the L2 is constructed on any of the mainstream public mainnets, individuals can stake not only bitcoin (wrapped BTC) \cite{wrappedtokens2019}, Ethereum/BNB, and stable coins such as USDT and USDC but also a variety of other assets to access services. However, there's a difference in the value of staked assets and the 'bandwidth' of services one can earn when using the native AI tokens compared to other cryptocurrencies. This difference is defined by a $q$ ratio. Typically, $q$ is valued at 0.1. So, for every one dollar's worth of AI tokens and other cryptocurrencies staked, the service volume ratio stands at 10:1. Both service providers and the global ledger adhere to this ratio. The rationale behind this design is to encourage more users to adopt the native tokens, promoting its market utility and commercialization.
\paragraph{The coordinator's aspect and system securities}
Coordinator nodes are responsible for directing messages and managing scheduling across the network, bridging the communication between clients and miners. While clients and miners interact seemingly directly, their exchanges are actually scheduled by these coordinators. Additionally, during each reward cycle, the coordinator nodes distribute system rewards to miners based on their respective contributions. Suppose that  \( R \) denotes the total rewards to be distributed in the given time span, \( C_k \) represents the contribution of the miner \( k \), and \( E \) is the overall count of miners. The aggregate contribution from all miners is represented by \( C_{\text{total}} \), calculated as \( C_{\text{total}} = \sum_{k=1}^{E} C_k \). Based on these parameters, the reward allocated to each individual miner \( k \), expressed as \( R_k \), is given by:
\begin{equation}
    R_k = \frac{C_k}{C_{\text{total}}} \times R
\end{equation}
\paragraph{Securities}
There are several security concerns with the current reward distribution protocol. Clients might unjustly give negative reviews to honest service providers. Malicious miners might either refuse to provide services or deliver incorrect service outputs, and it's impossible to prevent and verify this given the systme's distributed nature. In a more sophisticated attack, an entity might act as both a miner and a client to exploit the reward protocol. For example, they could flood the system with fake service exchange messages, artificially inflating a miner's contribution. The attackers can then get away with a significant portion of the system's periodic rewards, discouraging genuine miners.

Regarding malicious clients, the system can compare malicious reviews with other feedback. If a particular client's reviews consistently diverge from the majority, for example, if others frequently rate a service as ``GOOD'' while this client rates it as ``BAD'' or vice versa, access to the reputation protocol might be restricted for a set duration \(t_{\text{restricted}}\). If the client continues to submit biased reviews, the restriction period can increase exponentially: \(2t_{\text{restricted}}\), \(4t_{\text{restricted}}\), and so on, thus safeguarding the system against potential harm from the client's side.

Malicious miners can initiate denial of service attacks or provide low-quality services to a subset of clients. However, such actions will quickly lead to negative feedback on their reputation score. While reputation scores do not directly affect a miner's contribution and rewards, they can influence the likelihood of a miner being assigned a task. Since reputation scores can be publicly accessed from coordinator nodes, miners with poor reputations are less likely to be scheduled for a request or chosen by a client as a service provider.

Attacking the reward protocol involves an entity acting as both miners and clients, continually updating the system with false and non-existent service exchanges, maliciously building up its contribution over time. To counter this, two preventative measures are recommended. First, instead of allowing clients to choose the service provider directly, the coordinator nodes will select service providers based on their status and reputation. Nodes with higher reputation scores are more likely to be chosen. The random selection mechanism employed by the coordinator nodes can be inspired by the verifiable random function (VRF) from Chainlink \cite{chainlink2017}. To integrate a reputation-weighted random selection using VRF, one begins by generating an unpredictable and verifiable random number using VRF. This number is then used as input for a weighted random selection algorithm, where each of the $K$ miners has a selection weight determined by its reputation score. Techniques like the roulette wheel or stochastic acceptance can be utilized, ensuring that nodes with higher reputation scores have a higher probability of being selected. We call this function the weighted verifiable random function (AVRF). If we denote the selected miner index \(I_\text{request}\) as the routed miner for the request, then the AVRF can be written as:
\begin{equation}
I_\text{request} = \psi (\text{VRF},R)
\end{equation}
where \( R = (r_1, r_2, \ldots, r_K) \) represent the reputation scores of the \( K \) miners in the system, with \( r_i \) denoting the reputation score of the \(i^{th}\) miner.
Once clients are unaware of which miners might serve their requests, they lack the motivation to initiate the reward protocol attacks, as they might inadvertently boost the contributions of other miners. Moreover, the system can impose a threshold on the number of requests each client can make for specific AI services to prevent system flooding. Once this threshold is reached, the client will be temporarily frozen before it can send another request. 

\begin{algorithm}
\caption{Uncharged Protocol for Decentralized AI Services}
\begin{algorithmic}[1]

\Function{AcquireServicePass}{$\text{client\_token}$}
    \If{$\text{LockTokensIntoSmartContract(client\_token)}$}
        \State \Return $\text{GenerateServicePass()}$
    \Else
        \State \Return $\text{"Insufficient stake or token lock failed"}$
    \EndIf
\EndFunction

\Procedure{RequestService}{$\text{service\_pass}, \text{request}$}
    \If{$\text{IsValidServicePass(service\_pass)}$}
        \State $\text{provider} \gets \text{SelectServiceProvider()}$
        \If{$\text{provider.status} = \text{ready}$}
            \State $\text{service\_output} \gets \text{provider.Serve(request)}$
            \State $\text{feedback} \gets \text{GetClientFeedback(service\_output)}$
            \State \Call{ReviewService}{$\text{provider.id}, \text{feedback}$}
        \Else
            \State $\text{QueueRequest(request)}$
        \EndIf
    \Else
        \State \Return $\text{"Invalid Service Pass"}$
    \EndIf
\EndProcedure

\Function{SelectServiceProvider}{}
    \State $\text{vrf\_value} \gets \text{GenerateVRF()}$
    \State $\text{provider\_index} \gets \text{AVRF(vrf\_value, R)}$
    \State \Return $\text{provider\_list[provider\_index]}$
\EndFunction

\Function{GetClientFeedback}{$\text{service\_output}$}
    \State \Return $\text{ClientFeedback(service\_output)}$
\EndFunction

\Procedure{ReviewService}{$\text{provider\_id}, \text{feedback}$}
    \State $\text{ledger.UpdateReputation(provider\_id, feedback)}$
\EndProcedure

\end{algorithmic}
\end{algorithm}
\paragraph{Economics}
With the Uncharged Protocol, all token holders will be contributing to this system via a portion of the 5-10\% annual token inflation. Specifically, those who wish to access services must lock up their tokens, rendering them unable to sell these tokens until they finish using the service. Clients requiring long-term or continuous services may lock up their tokens for an indefinite period.

As the demand for services increases, leading to more tokens being locked up at a rate higher than the inflation rate due to the platform's growing market and commercial scale, the token economy undergoes an effective monetary deflation. This deflationary trend increases the value of tokens earned by service providers, encouraging them to offer a broader range of superior services.

Should there be a substantial decrease in service demand, the released tokens might flood the market, leading to an effective price drop beyond the standard inflation rate. This means the value of tokens may decrease, and the quality or quantity of services that providers can afford to offer might decline. However, due to the reduced demand, providers could choose to scale down their service offerings, thus reducing operational costs. Alternatively, adjustments could be made to the staking mechanism, recalibrating the number of tokens a client needs to stake to access a service.

Ultimately, clients in need of services fund the ecosystem via the time-value of their staked tokens. This ensures a smooth user experience with no micropayments, no transactional hurdles, and no unexpected fees.

\subsubsection{Charged approaches for decentralized AI services} Compared to uncharged approaches, charged approaches are more straightforward. The major process is that clients compensate service providers through subscription fees. These can be achieved per service request or as a single payment covering monthly or yearly durations. This approach is closely tied to the specific service being provided, as clients usually pay for a singular type of service to access. Although this design is simpler at the system level, with fewer security and economic complexities, there's an important responsibility for the platform: guiding clients to reputable service providers. Some providers might maliciously take clients' funds and later come back online with a new identity to commit fraud repeatedly. Occasional outages, even if unintended, can damage user experiences and decrease trust in the platform. As a result, it's crucial for the system to showcase trustworthy service providers prominently. Although some users might have specific preferences, the system should always highlight reliable service sources. Table. \ref{table_serviceFactors} outlines the factors the platform considers when listing service providers, with varying importance depending on the situation.

\begin{table}[!h]
\caption{Factors in Service Provider Display}%%%Table caption goes here
\label{table_serviceFactors}
\centering
\begin{tabular}{|c||c|}
\hline
\textbf{Factors} & \textbf{Data Structure Types} \\ %%%% Table header
\hline
Customer Rating of Service & float (typically 1-5/1-10) \\ %%%% Table body
\hline
Total Number of Subscribers & integer (0 to max clients) \\%%%% Table body
\hline
Total Number of Tokens Staked & float value $s$ \\
\hline
Types of AI Services Provided & string list $T$ \\
\hline
\end{tabular}
\end{table}%%%End of the table
Customer rating of service is typically represented as a float, indicating a one-star to five-star rating. The types of AI services provided include a string list that indicates the AI services the service provider supports. Like the uncharged approaches, a reputational protocol is also involved. Users are required to rate their service providers by submitting signed ratings to the global ledger. The global ledger then updates the miner's reputation based on these new ratings. Additionally, the coordinator node receives a portion of the payment as transaction fees. These fees typically cover the coordinator's operational costs and rewards but are kept moderate to prevent transaction friction and discourage miners. This fee structure also helps prevent the reward cheating attack mentioned earlier.

\subsection{Protocol implementations: network participants and process flow}
We focus on the operations carried out by various participants: clients, network coordinators, and miners.
We illustrate the process flow of different algorithms. To ensure the protocol accommodates as many types of AI services as possible, we have integrated both charged and uncharged economic approaches. Reputation protocols include rating scores from both sets of clients. This protocol is designed to enable large-scale commercialization.

\paragraph{The global ledger, coordinator nodes and peer to peer connections}
In our decentralized AI service network, the Global Ledger $\mathcal{L}$ plays a key role as a system record, logging all essential network interactions. The ledger contains three key components: the \textit{orders record}, the \textit{task cycle data}, and the \textit{node info}. The \textit{orders record} logs all orders placed by clients within the network, each containing the specific task details requested by a client; this includes the required service, data, and associated rewards in the charged case. The \textit{task cycle data} records the metadata of tasks that have undergone the full cycle of AI service provision and reward distribution in the network; it includes service generation signatures, related value exchanges, and potential comments and ratings. The \textit{node info} section saves the details of all registered service providers within the network, including their reputation and performance history. Collectively, these components of the ledger boost the network's performance by ensuring all operations are traceable and accessible in a timely manner. The \textit{coordinator node}, with the responsibility of publishing multi-signature transactions on the blockchain and updating contract states, plays a central role in managing ledger data and global states. Through the application of Byzantine Fault Tolerance (BFT) consensus, such as the Practical Byzantine Fault Tolerance (PBFT) algorithm \cite{PBFT}, it effectively maintains, updates, and synchronizes the Global Ledger $\mathcal{L}$. Besides storing and managing a synchronized copy of the global ledger, the coordinator nodes also act as data access points for other network participants. They provide on-demand access to the global ledger, ensuring its data is always available for different network operations.

Miners in the network are responsible for providing reliable data transfer links to supply AI service outputs. This data must be consistently accessible throughout the task cycle. Failure to do so can lead to filled orders receiving negative comments. It is also the client's responsibility to download the necessary data to their local storage for efficient service exchange processes.

\paragraph{Client cycle}
We provide an overview of the client cycle, which primarily includes the \texttt{Put}, \texttt{Get}, and \texttt{Rate} protocols. This offers a complete cycle from initiating the request to receiving a response and subsequently commenting on the services.
\begin{enumerate}
    \item \textbf{Put}: \textit{The client put orders requesting AI service}. 
Clients can request AI services from the network using staked pass or utility tokens. A client initiates the \texttt{Put} process by submitting an order to the network. Subsequently, coordinators have the authority to decide which service provider will handle the order, unless it's pre-specified in the charged case. Coordinator nodes submit job allocation messages to the global ledger. Clients can choose between free or charged services by providing relevant information in the appropriate sections of the order message. The selection of a paid service might result in higher quality service outputs.
\item  \textbf{Get}: \textit{Client retrieves model from the network}.
Clients can access the AI service output from the network as soon as it's complete. A direct peer-to-peer link connects the client and miner, initiated when the job allocation is logged in the network by coordinator nodes. The service provider then sends the AI results directly to the client. Once done, a confirmation is sent to the coordinator, which updates the global ledger with a completed job task. It is the miners' responsibility to ensure that their outputs are always made available to the clients to avoid negative effects on their reputations in the network.
\item \textbf{Rate}: Ratings are important components in the network, providing global ledger with the service qualitiy feedback of different service providers. Clients can initiate \texttt{Rate} by sending a signed message to the global ledger commenting on the latest service output of a service provider. \textit{Note:} the reputation score of a certain client on a service provider is always based on the latest ratings, meaning their previous comments are deleted and updated with the latest one.
\end{enumerate}
\paragraph{Mining Cycle (for service providers)}
We give an overview of the mining cycle of service providers. Service providers earn rewards by competing to earn higher reputation score in the network. 
\begin{enumerate}
    \item \textbf{Register}: Service Providers pledge their computational resources to the network. This is done by depositing collateral, via a transaction in the network, using \texttt{Miner.RegisterResource}. This collateral is locked in for the time intended to provide the service, and is returned upon request of the service provider if the provider decides to stop committing to the network, using \texttt{Miner.UnRegisterResource}. Once the service provider is registered, they can start generating model claims which will be added to the global ledger.
    
    \vspace{5pt}
\noindent$
\left[
\text{
\begin{tabular}{l}
\texttt{Miner.RegisterResource/UnRegisterResource}\\
$\bullet$ \texttt{INPUTS:}\\
~~ -- current global ledger $\mathcal{L}_t$\\
~~ -- registration request \texttt{register} \\
$\bullet$ \texttt{OUTPUTS:} current global ledger $\mathcal{L}_{t^{'}}$\\
\end{tabular}
}
\right.
$
\vspace{5pt}
\item \textbf{Service Executables}: After registration, the service providers execute various AI services offered by the network infrastructure, primarily determined by their capability to manage different services. The greater their computing power and bandwidth, the more services they can offer. The network schedules the distribution of task requests, and miners can accept incoming tasks once their service is online. After generating the output, the miner sends it back to the client and notifies the network that the specific request has been handled, thereby claiming their contribution.

    \vspace{5pt}
\noindent$
\left[
\text{
\begin{tabular}{l}
\texttt{Miner.Exec}\\
$\bullet$ \texttt{INPUTS:}\\
~~ -- current orders from global ledger $\mathcal{L}_t$\\
$\bullet$ \texttt{OUTPUTS:} signed message \texttt{sClaim} \\
\end{tabular}
}
\right.
$
 \vspace{5pt}
\item \textbf{Sending Outputs}: Service Providers are responsible for ensuring the availability of the generated output for a client $g_c$ throughout the full mining cycle. This is done through the \texttt{Miner.SendOutput} function. If a service provider fails to maintain the availability of these data, the network may invalidate the service, which will result in the service provider not receiving the contribution rewards. However, if a miner claims a contribution but doesn't provide the output to the client, it may lead to negative reviews, thus affecting the miner's operation of services.

    \vspace{5pt}
\noindent$
\left[
\text{
\begin{tabular}{l}
\texttt{Miner.SendOutput}\\
$\bullet$ \texttt{INPUTS:}\\
~~ -- order ID \texttt{oID}\\
~~ -- generated output $g_c$\\
$\bullet$ \texttt{OUTPUTS:} success status \texttt{sStatus}\\
\end{tabular}
}
\right.
$
\end{enumerate}
\begin{figure*}[!t] % the * makes it span both columns
    \centering
    \includegraphics[width=\textwidth]{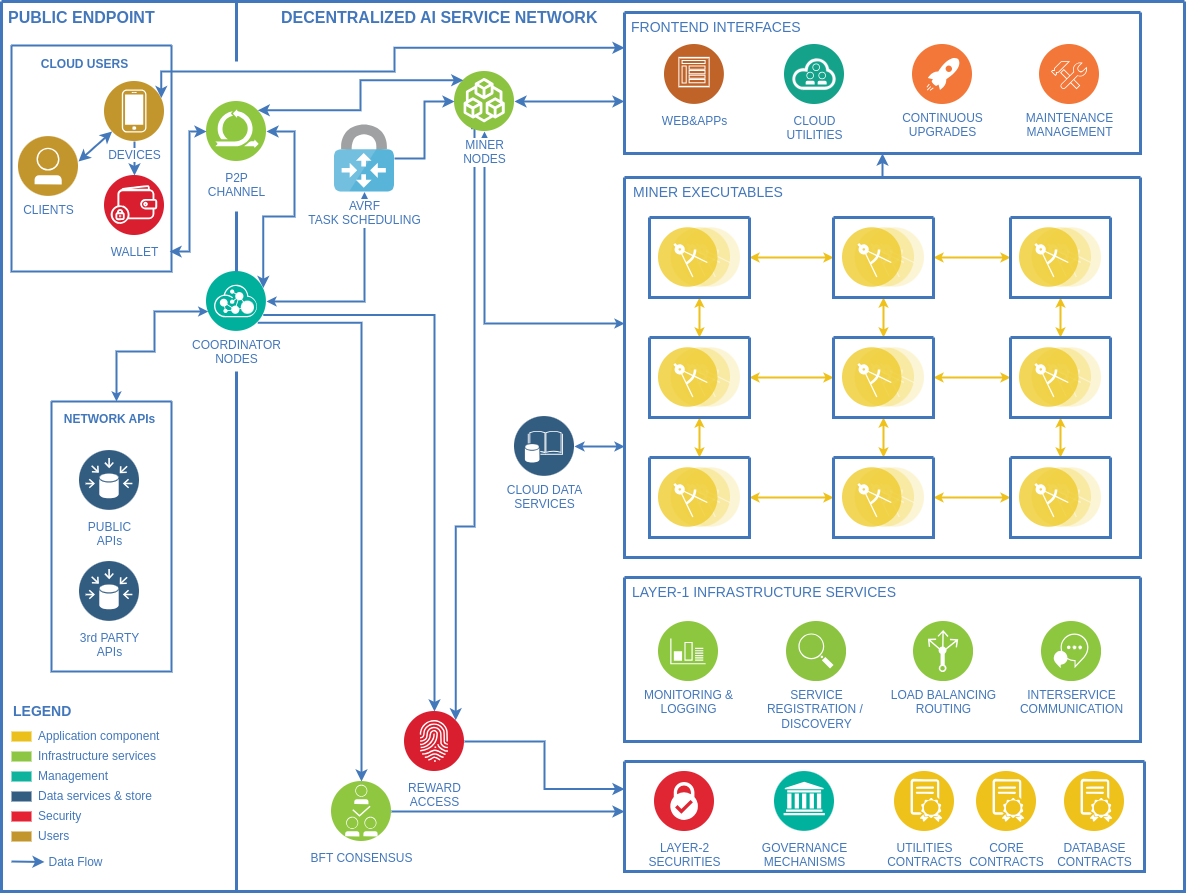}
    \caption{\textbf{Overview of the Decentralized AI Service Network:} This schematic represents the interconnected components of a decentralized AI service system. \textbf{On the left}, we have the public endpoint catering to users and devices, managed through wallets and coordinated via P2P channels and task scheduling. There are also network APIs for the convenience of developers and product development. \textbf{On the right}, frontend interfaces serve as the communication bridge between clients and the network, enabling easier integration for the community based on decentralized AI functionalities. The core of service in the network includes miner nodes and executables, which perform computations and return results. This entire ecosystem is supported by various infrastructure services, such as cloud data services, L1-L2 infrastructure and APIs.  Layer-1 focuses on primary infrastructure services, with distinct utilities, while Layer-2 emphasizes securities and governance. All components work interactively and cooperatively, ensuring efficient data flow, task distribution, and reward mechanisms.}
    \label{fig:overview_decentralized_ai}
\end{figure*}
\subsection{Discussions}
\subsubsection{Capability and scalability}
The system's transaction throughput performance, or in other words, the amount of information the system can process per second, is determined by its underlying design structure. The coordinator nodes act as a platform within the system, coordinating between clients and miners and enabling self-governance to initiate, process, and finalize services. Although the actual influx of transactions (including orders, confirmations, and ratings) will largely depend on the customer base and the total hash power of the network, the processing capability of the global states maintained by the coordinator nodes can be analyzed. Given the sizes of the orders, confirmations, and ratings messages, it can be estimated that a global ledger maintained by 50 coordinator nodes distributed worldwide can synchronize approximately 1000 full task cycles per second, as evidenced by the real L1-L2 network results in \cite{li2023proof}.

A significant advantage of the design lies in the allocation of computation-intensive tasks and storage to network participants. This strategy avoids overconsuming global storage and bandwidth, which could become costly, especially since updating global states is a synchronous process. The global ledger only stores information about orders, confirmations, and ratings, each of which is measured in kilobytes. Moreover, processing this information requires a computational complexity of $\mathcal{O}(1)$. Such a design enables the system to handle a virtually limitless number of task requests and service finalizations concurrently.
\subsubsection{Sercurity}
In most web3 protocols, the security of a protocol is guaranteed by economic incentives, i.e., attacking the system is more costly than complying with it. Similarly, in the designed platform, one would need to obtain more tokens than the counterparties to initiate attacks, which can often prove quite expensive. Unless the potential rewards are substantial, there is little incentive for someone to attack the protocol. 

In a scenario where the attack comes from the coordinator's side, it involves tampering with the rewarding process in the coordinator nodes' global ledger. This allows hackers to withdraw all tokens from the rewards distribution contract. To compromise the multi-sig design of the L1-L2 infrastructure, the attackers would need a ($m/c$) portion of the total staked tokens by the coordinator nodes. We call this \textit{Linear staking impact}, meaning that to be successful, an attacker must have a budget $B$ greater than a ($m/c$) portion the combined staked tokens of all coordinator nodes. More precisely, we mean that as a function of $m$, $B(m) = dm$ in a
network of $c$ coordinator nodes, each with a fixed staked amount $d$. Given our requirement for coordinator nodes to stake a significant amount of tokens to act as network coordinators, a hacker would need at least 10\% of the total circulation if 20\% of tokens are held by the honest coordinator nodes (assuming $m=18$ and $c=30$). Therefore, the cost of such an attack is generally much higher than the tokens in the reward contract.

In a scenario where the attack originates from the client's side, it involves flooding the system with requests using multiple fake identities, thereby halting the system by consuming all its resources. As discussed in the design sections, this is prevented by WRR, where a fake identity will be served only once or twice in a time frame while many others are being served. To effectively flood the system, they would need to increase their tokens, incurring a much higher cost. Another common attack pattern involves giving malicious reviews to honest miners during a service. However, if a client's ratings deviate significantly from the majority of reviews most of the time, the client might be suspended  from commenting for a period of time by the coordinator nodes to mitigate its negative influence.

In scenarios where the attacker originates from the miner's side, it typically involves miners attempting to maliciously increase their contribution, as this directly correlates to the distribution of rewards for uncharged services. This contribution can be built up through two variables: the service weight (determined directly by the community DAO) and the number of services provided by the miner. While it is extremely difficult to manipulate the service weight, miners may be incentivized to exaggerate the number of services they provide. Fake clients might artificially inflate the number of services, but this is countered by the AVRF scheduler and a threshold for the number of service accesses in a given timeframe. Moreover, fake requests will inadvertently boost the contributions of others in the current protocol, thereby offsetting the negative impact. A miner might also try to execute a denial of service attack, but their reputation would rapidly decline due to subpar service quality. Additionally, community members can vote out malicious miners via the DAO. But most importantly, platform developers should design an intricate recommendation and list algorithm that prioritizes statistically reliable and honest service providers on the frontend (be it a website or app). This ensures that users are more likely to select top-tier service providers, as these are presented with priority. As a result, if these prioritized providers offer charged services, they are more likely to be trusted by clients to be committed to the service deal, rather than "rugging" once they receive payment.

In general, in a staking-intensive web3 environment, many attack types can be mitigated by adjusting various staking protocols and the time required for the staking and unstaking processes. To the best of our knowledge, the system is robust against different types of attacks. The fundamental idea is that we aim to keep the cost of attacking the system high at all times, regardless of the angle from which the attack may originate.
\subsubsection{Advantages}
We believe one of the major advantages of the solution lies in its consensus mechanism design. This provides significant capacity and scalability benefits compared to other solutions in this field. Generally, Web3 infrastructures are less efficient due to their distributed nature and inherent lack of trust. However, with this design, the network coordinator, which maintains the global ledger and global states, is relieved from handling large data storage or the heavy computation tasks common in most AI servicing processes. Instead, these tasks are delegated to participant nodes with ample resources. Participants are given strong cryptoeconomic incentives to act honestly and diligently. This creates a system that is largely self-governing, further enhancing the solution's capacity and scalability. Beyond the reputation protocol, participants are regulated by a community DAO. For instance, if any service provider fails to provide the necessary service and bandwidth for a smooth exchange process, they may face penalties. This could come in the form of other nodes on the network voting against them in a community motion within the DAO. Consequently, the consensus mechanism ensures that participants remain committed to their orders and services, guaranteeing system liveliness.

Another major advantage of the solution, compared to others, is the design of its L1-L2 system structure, which ensures easy system upgradability. AI is a rapidly changing industry, with new types of services emerging daily. The protocol employs Layer-2 (on-chain) applications for depositing, withdrawing, and transferring users' assets, while the majority of operations are conducted on Layer-1 (off-chain) to facilitate upgradability. To integrate new services, miners and coordinators can simply upgrade to the latest version of the software. Subsequently, clients will have the ability to specify these new service types in their orders. In theory, the system can incorporate any kind of AI service into the L1 infrastructure.

\section{Conclusions and Future Works}
In this work, we have provided a comprehensive review of the opportunities and challenges related to merging web3 and the AI marketplace. We thoroughly studied the advantages of both fields and the challenges involved in their integration. We also presented our solutions on this topic, with a primary focus on the framework's commercial rationality, security, and efficiency. We believe that the framework should first demonstrate feasibility and the potential to catalyze large-scale commercialization before focusing on its security and efficiency. We began with the user experience in mind and then identified ways to technically realize our vision. In general, we've made contributions in two main areas: firstly, we offer an overview of the commercial landscape of web3 and the AI marketplace, highlighting both opportunities and challenges in this commercial avenue, and secondly, we proposed our solutions.

To our knowledge, our platform, which supports both charged and uncharged AI services, is the first in the industry to introduce such a framework with the key protocols presented. It emphasizes user experience, maximizing its potential for widespread adoption, yet it is intricately designed to ensure the platform's resilience against the various types of attacks prevalent in the web3 industry. While the web3 ecosystem is occasionally perceived as less efficient by the web2 community, our system's total throughput demonstrates potential in serving customers worldwide. We showed that with a optimized design structure, high-efficiency web3 platforms can be realized without compromising their distributed nature, thus ensuring broader accessibility. In summary, we have proposed a solution that allows anyone holding cryptocurrencies to access a range of AI services, whether they seek free offerings or wish to pay for customized services.

One aspect not covered in this paper is the execution of experiments involving different sub-protocols within the designed framework, especially those concerning the interaction between clients and miners as actual tasks are resolved. This omission is primarily because any simulation in this regard would merely represent a specific case of the system's capacity and throughput. However, analyzing the protocol from a financial perspective is set as part of our future work. We aim to engage the current crypto mining infrastructure in the web3 community by introducing network utility tokens and implementing a comprehensive version of the framework with detailed parameters. This would allow for a thorough analysis of the system's performance on real-world tasks, paving the way for further developments and deeper understanding of the AI marketplace within web3.

\bibliographystyle{unsrtnat}
\bibliography{references}  %%% Uncomment this line and comment out the ``thebibliography'' section below to use the external .bib file (using bibtex) .




% 	\bibitem{keshet2016prediction}
% 	Keshet, Renato, Alina Maor, and George Kour.
% 	\newblock Prediction-Based, Prioritized Market-Share Insight Extraction.
% 	\newblock In {\em Advanced Data Mining and Applications (ADMA), 2016 12th International 
%                       Conference of}, pages 81--94,2016.

% \end{thebibliography}

\end{document}